\documentclass{article}

\PassOptionsToPackage{numbers, sort&compress}{natbib}

\usepackage[final]{neurips_2020_ml4ps}

\usepackage[T1]{fontenc}
\usepackage[utf8]{inputenc}
\usepackage{color,xcolor}
\definecolor{dark-grey}{rgb}{0.6, 0.6, 0.6}
\definecolor{dark-red}{rgb}{0.7, 0.0, 0.25}

\usepackage{hyperref}
\usepackage{amsmath, amsfonts, dsfont, mathtools, amssymb}
\usepackage{enumitem} 
\usepackage{xspace} 
\usepackage{subcaption}
\usepackage{microtype}
\usepackage{algorithm, algpseudocode}
\usepackage{url}
\usepackage{xfrac}
\usepackage{booktabs}
\usepackage{wrapfig}
\usepackage{multirow}
\usepackage{graphicx, graphbox}
\usepackage[font=footnotesize]{caption}

\newcommand{\toolfont}[1]{\textsc{#1}}

\newcommand{\eg}{{e.\,g.}~}
\newcommand{\ie}{{i.\,e.}~}

\newcommand{\mf}{\text{$\mathcal{M}$\kern-0.1pt-flow}}
\newcommand{\mfe}{\text{$\mathcal{M}_{\kern-0.2pte\kern-0.3pt}$-flow}}

\newcount\Comments  
\definecolor{darkgreen}{rgb}{0,0.5,0}
\definecolor{darkred}{rgb}{0.7,0,0}
\definecolor{teal}{rgb}{0.1,0.6,0.7}
\definecolor{blue}{rgb}{0.0,0.1,0.9}
\definecolor{orange}{rgb}{1.,0.7,0.0}
\definecolor{lightblue}{rgb}{0.70, 0.80, 0.89}
\newcommand{\kibitz}[2]{\ifnum\Comments=1{{\textcolor{#1}{\textsf{\footnotesize [#2]}}}}\fi}

\DeclareMathOperator*{\argmax}{arg\,max}

\title{Hierarchical clustering in particle physics\\through reinforcement learning}

\author{%
    Johann Brehmer\\%
    New York University\\%
    \texttt{johann.brehmer@nyu.edu}
    \And%
    Sebastian Macaluso\\
    New York University\\
    \texttt{sm4511@nyu.edu}
    \AND%
    Duccio Pappadopulo\\
    Bloomberg\thanks{Work done before joining Bloomberg.}\\
    \texttt{dpappadopulo@bloomberg.net}
    \And%
    Kyle Cranmer\\
    New York University\\
    \texttt{kyle.cranmer@nyu.edu}
}

\begin{document}

\maketitle

\begin{abstract}
    Particle physics experiments often require the reconstruction of  decay patterns through a hierarchical clustering of the observed final-state particles. We show that this task can be phrased as a Markov Decision Process and adapt reinforcement learning algorithms to solve it. In particular, we show that Monte-Carlo Tree Search guided by a neural policy can construct high-quality hierarchical clusterings and outperform established greedy and beam search baselines.
\end{abstract}

\section{Introduction}

Particle interactions in collider experiments often produce highly energetic quarks and gluons (or \emph{partons}). These elementary particles then radiate more and more quarks and gluons in a series of successive binary splittings, ultimately leading to a \emph{jet}: a spray of stable particles that can be measured in a detector. This \emph{parton shower} process follows the laws of quantum chromodynamics, our understanding of it is encoded in sophisticated simulators.

Reconstructing the properties of the original elementary particles from the observed final-state particles is an important step in the data analysis in particle physics experiments, including those at the Large Hadron Collider at CERN. It is often crucial in the search for new particles or the measurements of the properties of the Higgs boson. This \emph{jet clustering} problem can be phrased as an inference task that aims to invert the generative radiation process. Given a set of observed final-state particles or leaf nodes, the goal is to construct the most plausible binary tree of particle splittings.

Unfortunately, for more than a few final-state particles, the search space of this combinatorial optimization problem is too large to find the true maximum-likelihood tree. The industry standard is to solve this problem with a greedy algorithm based on one of several heuristics (for instance the popular $k_T$, Cambridge-Aachen, and anti-$k_T$ algorithms~\cite{Catani:1993hr, Ellis:1993tq, Dokshitzer:1997in, Cacciari:2008gp}). Improvements may come from questioning two aspects of this industry standard. First, we may be able to improve clustering by switching from optimizing heuristics to maximizing the likelihood~\cite{TreeAlgorithms}. Second, there are more powerful algorithms than the greedy optimization, for instance beam search, probabilistic methods~\cite{Ellis:2012sn}, and simulation-based inference methods~\cite{Cranmer:2019eaq}.

In this paper we propose to phrase this combinatorial optimization problem as a Markov Decision Process (MDP), which allows us to use reinforcement learning (RL) methods to solve it. In particular, we adapt Monte-Carlo Tree Search (MCTS) guided by a neural network policy to the problem of jet clustering. This approach closely follows the AlphaZero algorithm~\cite{silver2016mastering, silver2017mastering, silver2017mastering2}, which achieved superhuman performance in a range of board games, demonstrating its ability to efficiently search large combinatorial spaces. We also test imitation learning, specifically Behavioral Cloning, and train a policy to imitate the actions that reproduce the true tree from the generative model. While (model-free) RL methods have been used in the context of jet grooming, \ie pruning an existing tree to remove certain backgrounds~\cite{Carrazza:2019efs}, they have not yet been used for clustering, that is, the construction of the binary tree itself. In first experiments we demonstrate that the MCTS clustering agent outperforms not only the greedy industry standard, but also beam search. Finally, we discuss which steps need to be taken to scale this approach to real-life applications. The code used for this study is available at \url{https://github.com/johannbrehmer/ginkgo-rl}.

\section{Jet clustering as a Markov Decision Process}

\paragraph{Problem statement.} We describe the parton shower as a stochastic generative process in which a set of particles described by their four-momenta at time step $t$, $z_t = \{p_{t,1}, p_{t,1}, \dots, p_{t, n_t}\}$, undergoes successive binary splittings $z_t \to z_{t+1} \sim p_s(z_{t+1} | z_t)$. The successor state $z_{t+1}$ contains $n_t - 1$ unchanged particles from $z_t$ as well as the children of one radiating particle. The whole decay process thus forms a binary tree. The individual splittings are Markov, and their probability densities $p_s(z_{t+1} | z_t)$ are known and tractable. In particular, each splitting has to satisfy the energy-momentum conservation law: a splitting $p_{t, i} \to p_{t+1, i}, p_{t+1, j}$ satisfies $p_{t, i} = p_{t+1, i} + p_{t+1, j}$. The initial state is a single elementary particle, $z_1 = \{p_{1, 1}\}$. The process terminates after $N$ splittings, after which the final-state momenta $x = z_N$ are observed. For a detailed description of the parton shower, see \eg Refs.~\cite{Ellis:1991qj, Buckley:2011ms}.

Given an observed final state $x = z_N = \{p_{N,1}, p_{N,2}, \dots, p_{N,n_N} \}$, we study the problem of inferring the maximum-likelihood latent binary tree $z^* = \argmax_{\{z_1, \dots, z_N\}} p(x | \{z_1, \dots, z_N\})$ with
\begin{equation}
    p(x | \{z_1, \dots, z_N\}) = p_s(x | z_{N - 1})\prod_{t=1}^{N - 2} p_s(z_{t+1} | z_t) \,.
    \label{eq:likelihood}
\end{equation}

\paragraph{Markov Decision Process (MDP).} We treat the problem of clustering as an MDP $(\mathcal{S}, \mathcal{A}, P, R)$:
\begin{itemize}
    \item The state space $\mathcal{S}$ is given by all possible particle sets at any given point during the clustering process, $s = z_t$.
    \item The actions $\mathcal{A}$ are the choice of two particles $a=(i, j)$ with $1 \leq i < j \leq n_t$ to be merged.
    \item The state transitions $P$ are deterministic and update $z_t$ to $z_{t - 1}$ by replacing the particles $p_{t,i}$ and $p_{t,j}$ with a parent $p_{t-1,i} = p_{t,i} + p_{t,j}$. All other particles are left unchanged, each state transition thus reduces the number of particles by one.
    \item The rewards $R$ are the splitting probabilities, $R(s=z_t, a=(i,j)) = \log p_s(z_{t} | z_{t-1}(i,j))$.
    \item The MDP is episodic and terminates when only a single particle is left.
\end{itemize}

\begin{wrapfigure}{R}{0.45\textwidth}
    \vskip-0.45cm
    \centering
    \includegraphics[width=\linewidth,clip,trim=0.4cm 0.6cm 0.4cm 0.4cm]{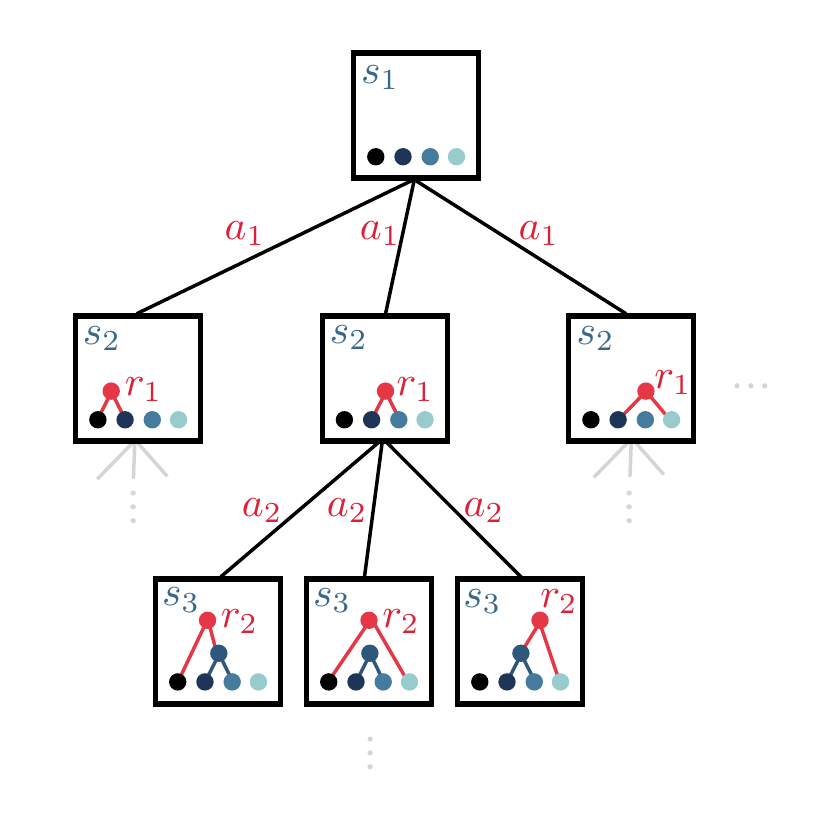}%
    \vskip-0.1cm
    \caption{Jet clustering as a Markov Decision Process. States $s$ (squares) represent (partial) clusterings of the original particles (small circles), the agent begins in the unclustered state (top square). Each action $a$ chooses a pair of particles in the current state (which may be either part of the original particle set or the result of a partial clustering) to be merged next. The reward $r$ is the log likelihood of the corresponding $1 \to 2$ splitting.}
    \label{fig:mdp_sketch}
    \vskip-0.4cm
\end{wrapfigure}

An agent solves the jet clustering problem by first considering the state of all observed, final-state particles and choosing which two to merge into a parent. It receives the log likelihood of this splitting as reward. Next, it considers the reduced set of particles where the two chosen particles have been replaced by their proposed parent, chooses the next pair of particles to merge, and so on. Rolling out an episode leads to a proposed clustering tree $z = \{z_1, \dots, z_N\}$, with the total received reward being equal to the log likelihood of this tree following Eq.~\eqref{eq:likelihood}. We illustrate this setup in Fig.~\ref{fig:mdp_sketch}.

\paragraph{Limitations.} Our description of the parton shower as a binary process and in particular the assumption that the final state is perfectly observable are approximations to the true physical process. In reality, the final states of the shower first have to form stable particles (hadrons) and then interact with a complicated detector to be measured. The clustering problem is thus actually a partially observable MDP. These effects, which are also not modeled in state-of-the-art clustering algorithms, go beyond the scope of this proof-of-concept paper.

\clearpage

\section{Algorithms}

\paragraph{Monte-Carlo Tree Search (MCTS).} The formulation of jet clustering as an MDP allows us to use any (model-free) reinforcement learning (RL) algorithm to tackle it. Since the state transition model is known (and deterministic), we instead use a model-based planning approach to leverage this knowledge. We choose MCTS~\cite{silver2016mastering}, which builds a search tree over possible clusterings $z$ by rolling out a number of clusterings. During these roll-outs, at each state $s$ we choose the action $a$ that maximizes the upper confidence bound on the action values (PUCT)
\begin{equation}
    U_{s, a} = Q_{s, a} + c \, \pi(s, a) \, \frac {\sqrt{N_s}} {1 + N_{s, a}}\,.
\end{equation}
Here $\pi(s, a)$ is a learnably policy implemented as a neural network, $Q_{s, a}$ is the mean normalized reward received after chosing action $a$ in state $s$, $N_s$ ($N_{s,a}$) is the number of times a state $s$ has been visited (and action $a$ has been chosen), and $c$ is a hyperparameter that balances exploration and exploitation. Since the episode length is given by the number of leaf particles, we roll out all MCTS trajectories until termination.

After a fixed number of MCTS roll-outs, the agent ultimately picks the action $a^*$ that lead to the largest individual roll-out reward. The policy $\pi(s, a)$ is trained to agree with the MCTS decisions by maximizing the log likelihood $\log \pi(s, a^*)$.

While the MCTS algorithm should be able to learn good policies from raw data, we find that with limited training time it benefits from feature engineering and a suitable initialization. As inputs into the neural network $\pi(s,a)$ we use not only the four-momenta of all current particles, but also the splitting probabilities $p_s(z_{t+1} | z_t(i, j))$ for all possible actions $a = (i,j)$. In addition, we initialize the MCTS search tree at each step by running beam search with a small beam size $b$.

\paragraph{Behavioral Cloning (BC).} Next, we consider a clustering algorithm based on imitation learning, specifically Behavioral Cloning: a policy $\pi$ is trained to imitate the actions that reconstruct the true trees, which we can extract from the generative model, by maximizing $\log \pi(s, a_\mathrm{truth})$.

\paragraph{MLE-BC.} For samples with a small number of final-state particles $n_N$ we can also construct the exact maximum-likelihood tree using the algorithm from Ref.~\cite{greenberg2020compact}, and train a policy to imitate it. The MLE tree becomes impractically slow for $n_N > 10$; for these samples we continue to use the simulator truth trees as demonstrator actions. 

\paragraph{BC-MCTS.} Finally, we consider an MCTS planner where the policy $\pi$ is pretrained by BC, again using the true decay trees as demonstrator actions.

\section{Experiments}

\paragraph{The \toolfont{Ginkgo} simulator.} We demonstrate the MCTS clustering algorithm in a simplified setup where the likelihood of each individual tree is tractable. To this end we use \toolfont{Ginkgo}~\cite{ToyJetsShowerPackage}, a toy generative model for jet physics that captures essential aspects of realistic simulators of the parton shower. We implement the MDP described above based on this simulator and provide a standard \toolfont{OpenAI~Gym}~\cite{brockman2016openai} interface.

\paragraph{Setup.} We repeatedly simulate the parton shower for a single initial elementary particle. The final states of these decays are then clustered by our algorithms. Neural policies are trained for 60\,000 steps (MCTS) or $10^6$ steps (BC). The clustered trees are evaluated on 500 samples, using the mean log likelihood as given in Eq.~\eqref{eq:likelihood} as metric.

\begin{wraptable}{R}{0.51\textwidth}
    \centering
    \footnotesize
    \begin{tabular}{l@{}r}
        \toprule
        Algorithm & Log likelihood \\
        \midrule
        Random & $-198.3_{\textcolor{dark-grey}{\pm 23.8}}$ \\
        Greedy & $ -96.0_{\textcolor{dark-grey}{\hphantom{0} \pm 0.0}}$ \\
        Beam search ($b=5$) & $ -94.1_{\textcolor{dark-grey}{\hphantom{0} \pm 0.0}}$ \\
        \hspace*{0.3cm}($b=20$) & $ -93.5_{\textcolor{dark-grey}{\hphantom{0} \pm 0.0}}$ \\
        \hspace*{0.3cm}($b=100$) & $ -93.3_{\textcolor{dark-grey}{\hphantom{0} \pm 0.0}}$ \\
        \hspace*{0.3cm}($b=1000$) & $ -93.3_{\textcolor{dark-grey}{\hphantom{0} \pm 0.0}}$ \\
        \midrule
        MCTS ($b=3$, $n_\mathrm{MCTS}=10$) & $ -93.4_{\textcolor{dark-grey}{\hphantom{0} \pm 0.1}}$ \\
        \hspace*{0.3cm}($b=5$, $n_\mathrm{MCTS}=20$) & $ -93.3_{\textcolor{dark-grey}{\hphantom{0} \pm 0.1}}$ \\
        \hspace*{0.3cm}($b=20$, $n_\mathrm{MCTS}=50$) & $ -93.0_{\textcolor{dark-grey}{\hphantom{0} \pm 0.1}}$\\
        \hspace*{0.3cm}($b=100$, $n_\mathrm{MCTS}=200$) & $ \mathbf{-92.8}_{\textcolor{dark-grey}{\hphantom{0} \pm 0.1}}$ \\
        BC & $-108.5_{\textcolor{dark-grey}{\hphantom{0} \pm 1.6}}$ \\
        MLE-BC & $-108.3_{\textcolor{dark-grey}{\hphantom{0} \pm 1.7}}$ \\
        BC-MCTS ($b=3$, $n_\mathrm{MCTS}=10$) &  $ -93.5_{\textcolor{dark-grey}{\hphantom{0} \pm 0.1}}$ \\
        \hspace*{0.3cm}($b=5$, $n_\mathrm{MCTS}=20$) & $ -93.3_{\textcolor{dark-grey}{\hphantom{0} \pm 0.1}}$ \\
        \hspace*{0.3cm}($b=20$, $n_\mathrm{MCTS}=50$) & $ -93.0_{\textcolor{dark-grey}{\hphantom{0} \pm 0.1}}$ \\
        \hspace*{0.3cm}($b=100$, $n_\mathrm{MCTS}=200$) & $ \mathbf{-92.8}_{\textcolor{dark-grey}{\hphantom{0} \pm 0.1}}$ \\
        \midrule
        \multicolumn{2}{l}{MCTS ($b=5$, $n_\mathrm{MCTS}=20$) variations:}\\
        \hspace*{0.3cm}exploit more ($c=0.1$) & $ -93.4_{\textcolor{dark-grey}{\hphantom{0} \pm 0.1}}$ \\
        \hspace*{0.3cm}explore more ($c=10$) & $ -93.3_{\textcolor{dark-grey}{\hphantom{0} \pm 0.1}}$ \\
        \hspace*{0.3cm}no $p_s$ as input to policy & $ -93.8_{\textcolor{dark-grey}{\hphantom{0} \pm 0.0}}$ \\
        \hspace*{0.3cm}final decision based on PUCT & $ -94.6_{\textcolor{dark-grey}{\hphantom{0} \pm 0.3}}$ \\
        \hspace*{0.3cm}no beam search roll-outs  & $ -97.3_{\textcolor{dark-grey}{\hphantom{0} \pm 1.3}}$ \\
        \hspace*{0.3cm}no PUCT roll-outs & $ -93.9_{\textcolor{dark-grey}{\hphantom{0} \pm 0.0}}$ \\
        \hspace*{0.3cm}instead of NN, use policy $\propto p_s$ & $ -93.5_{\textcolor{dark-grey}{\hphantom{0} \pm 0.0}}$ \\
        \hspace*{0.3cm}instead of NN, use random policy & $ -93.9_{\textcolor{dark-grey}{\hphantom{0} \pm 0.0}}$ \\
        \bottomrule
    \end{tabular}
    \caption{Mean log likelihood of clustered trees (larger is better, best results are bold). We show the mean and its standard error between five models trained with different random seeds. MCTS (middle) yields higher-quality hierarchical clusterings than the baselines (top).}%
    \label{tbl:results}%
    \vskip-0.6cm
\end{wraptable}
%
\paragraph{Baselines.} We compare MCTS (with $c=1$) and BC agents to a greedy algorithm that at each state picks the action with the maximum splitting likelihood $p_s$, a beam search algorithm that maintains the $b$ most likely clusterings while descending down the search tree, and a random policy. For jets with a small number of final-state particles $n_N$ we also compute the exact MLE tree following Ref.~\cite{greenberg2020compact}, though the $O(3^{n_N})$ complexity of that algorithm makes this intractable for large $n_N$.

\begin{figure}%
    \centering%
    \includegraphics[width=0.475\linewidth]{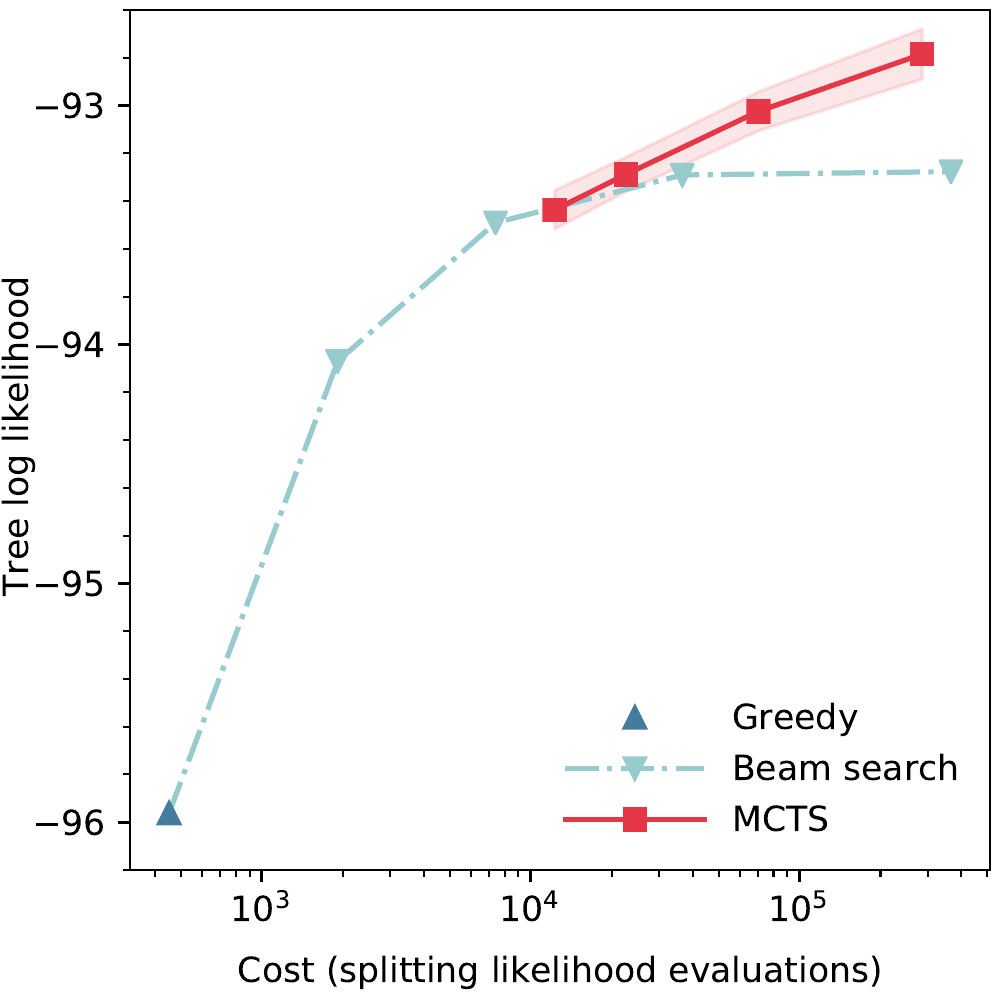}%
    \hspace*{0.04\linewidth}%
    \includegraphics[width=0.475\linewidth]{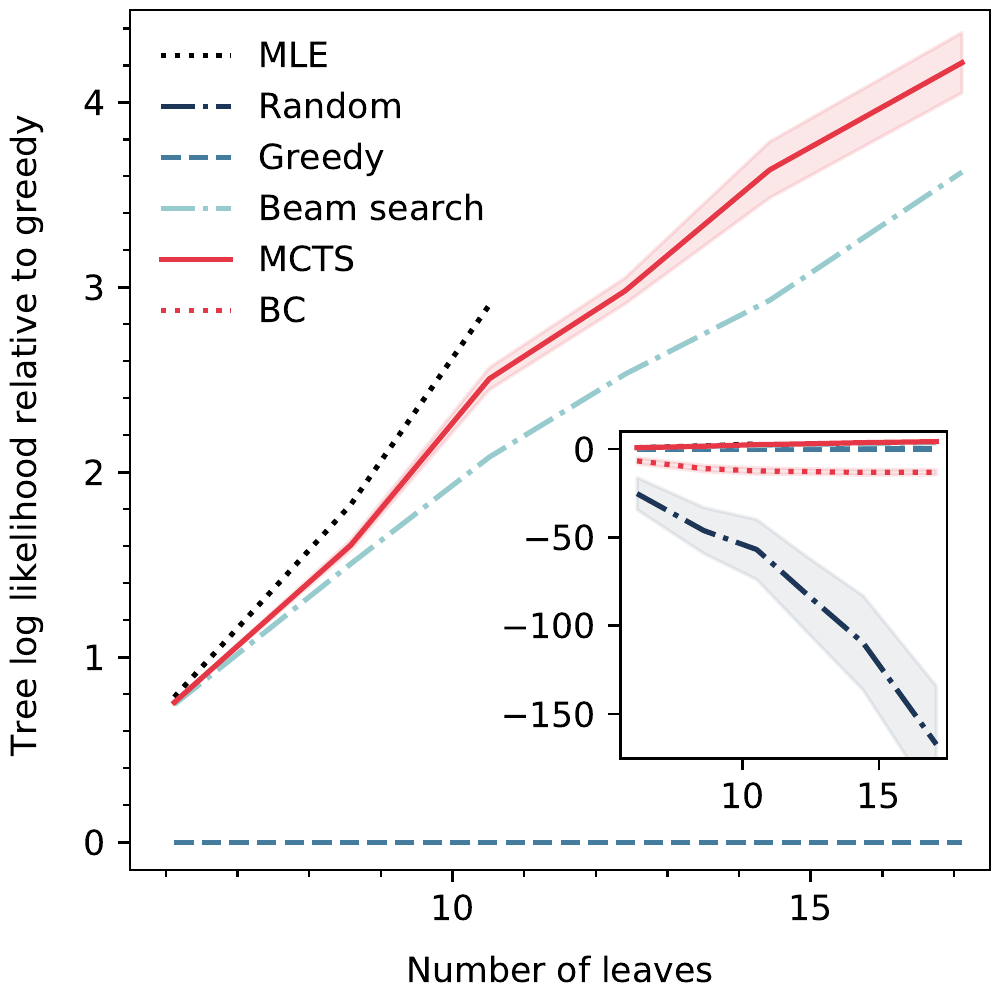}%
    \caption{Mean log likelihood of clustered trees (larger is better). We show the mean and its standard error between five models trained with different random seeds. \textbf{Left}: against the computational cost, measured as the number of evaluations of the splitting likelihood $p_s$ required by the different algorithms. For beam search and MCTS we show four different hyperparameter settings. \textbf{Right}: as a function of the number of final-state particles (leaves of the tree), using the best-performing (and most computationally expensive) hyperparameter setup for each algorithm. MCTS (solid, red) gives the highest-quality tree clusterings.}%
    \vskip-0.5cm
    \label{fig:results}%
\end{figure}

\paragraph{Results.} We show the results in Tbl.~\ref{tbl:results} and in Fig.~\ref{fig:results}, where we plot the clustering log likelihood against the computational cost of the clustering algorithms (left panel) and against the number of final-state particles (middle and right panel). While the greedy and beam search baselines lead to a robust performance at low computational cost, MCTS planning can generate hierarchical clusterings of a markedly higher likelihood. This advantage is more pronounced at larger number of final-state particles, showing that MCTS can explore large combinatorial spaces better than the baselines. We also observe that for small-to-medium trees (where maximum likelihood trees are tractable) the MCTS clusterings are close to optimal.

Imitation learning does not perform very well. While the BC and MLE-BC policies generate clusterings that are substantially better than random, they are still worse than the greedy baseline. Pretraining the policy that guides the MCTS planner on BC does not make a significant difference compared to the simple MCTS setup. Further ablation studies of our MCTS setup show that the neural network policy performs better than simply using a random policy or a policy that is proportional to the splitting likelihood $p_s$ to explore the search tree. Feeding the splitting likelihoods $p_s$ as additional features into the policy network also helps the models, and the initialization of the search tree by running beam search is particularly important for a good performance.

\section{Discussion}

We have presented new algorithms for a hierarchical clustering problem in particle physics based on Monte-Carlo Tree Search guided by a neural policy as well as on Behavioral Cloning. In a simplified scenario using the \toolfont{Ginkgo} simulator, the MCTS algorithm generated higher-quality hierarchical clusterings than greedy and beam search baselines.

Scaling this approach to real-life particle analyses requires a few more steps. First, we need to switch to a realistic model of the parton shower. This is straightforward for the RL algorithm itself. To compute the reward function, one could add the capability to evaluate the splitting likelihoods to existing simulators or switch to heuristics. Unlike greedy and beam search algorithms, the MCTS clustering algorithm can optimize hierarchical clusterings based on \emph{any} reward function, which may be delayed up to the termination of the episode. This allows us to optimize it directly based on some downstream task, letting us tailor clustering algorithms to a given problem.

A second necessary step is the careful analysis of infrared and collinear safety, invariance properties of the hierarchical clustering under certain additional splittings in the generative process. In addition, we will need to make sure that the algorithm performs reliably when presented with high-energy events, which are scarce in the training data, but of particular interest in many physics analyses. More broadly, we need to analyze how much higher-likelihood reconstructed trees can improve the results of downstream physics analyses.

Finally, the RL algorithms need to become faster to be useful under real-life conditions. Adding a value network instead of rolling out all trajectories until termination as well as using neural architectures and search trees that directly reflect the hierarchical clustering structure of the problem may help improve the performance.

\section*{Broader impact}

The application of reinforcement learning methods to hierarchical clusterings in particle physics has the potential to improve data analysis methods in collider experiments. In a simplified setup we demonstrated that it can outperform a greedy algorithm, which is the current industry standard in particle physics analyses. If this encouraging result persists under more realistic conditions and the algorithm can be made faster, our approach may make precision measurements of the Standard Model of Particle Physics, studies of properties of the Higgs boson, and searches for new particles more efficient, and ultimately help us understand the fundamental properties of the universe a little better. The RL approach may also improve the quality of hierarchical clusterings in other contexts, for instance in genomics~\cite{greenberg2020compact}.

\section*{Acknowledgements}

We would like to thank Craig S.\ Greenberg, Nicholas Monath, Ji-Ah Lee, Patrick Flaherty, Andrew McGregor, and Andrew McCallum for their collaboration on efficient maximum likelihood estimation on Ginkgo with a trellis. We are grateful to Matt Drnevic for his work on Ginkgo and want to thank Marcel Gutsche for great discussions. Our thanks also goes to the authors and maintainers of \toolfont{gym}~\cite{brockman2016openai}, \toolfont{Jupyter}~\cite{Kluyver2016JupyterN}, \toolfont{Matplotlib}~\cite{Hunter:2007}, \toolfont{NumPy}~\cite{numpy:2011}, \toolfont{PyTorch}~\cite{paszke2017automatic}, \toolfont{sacred}~\cite{klaus_greff-proc-scipy-2017}, and \toolfont{stable-baselines}~\cite{stable-baselines}. We are grateful for the support of the National Science Foundation under the awards ACI-1450310, OAC-1836650, and OAC-1841471, the Moore-Sloan data science environment at NYU, as well as the NYU IT High Performance Computing resources, services, and staff expertise.

\bibliography{references}

\end{document}